\documentclass[conference]{IEEEtran}
\IEEEoverridecommandlockouts
\usepackage{cite}
\usepackage{amsmath,amssymb,amsfonts}
\usepackage{algorithmic}
\usepackage{graphicx}
\usepackage{textcomp}
\usepackage{xcolor}
\usepackage{float}
\usepackage{multirow}
\usepackage{dblfloatfix}
\usepackage{microtype}
\usepackage{flushend}
\usepackage{xspace}
\usepackage{placeins}
\usepackage{comment}
\def\BibTeX{{\rm B\kern-.05em{\sc i\kern-.025em b}\kern-.08em
T\kern-.1667em\lower.7ex\hbox{E}\kern-.125emX}}

\newcommand{\ourwork}{LoCoMAP}

\usepackage{fancyhdr}
\fancypagestyle{arxivfirstpage}{\fancyhf{}\fancyfoot[C]{\parbox{\textwidth}{\fontsize{7}{8}\selectfont\raggedright \copyright{} 2026 IEEE. Personal use of this material is permitted. Permission from IEEE must be obtained for all other uses, in any current or future media, including reprinting/republishing this material for advertising or promotional purposes, creating new collective works, for resale or redistribution to servers or lists, or reuse of any copyrighted component of this work in other works.}}}

\begin{document}

\title{Topology Survives Shuffling: Robust Long-Context
Memory Access Prediction for Graph Analytics}

\author{
\IEEEauthorblockN{
Dongyan Sun\textsuperscript{*},
Neelesh Gupta\textsuperscript{*\dag},
Rajgopal Kannan\textsuperscript{\ddag}, Viktor K. Prasanna
}
\IEEEauthorblockA{
University of Southern California, USA\\
\textsuperscript{\ddag}DEVCOM Army Research Office, USA\\
Email: \{dongyans, neeleshg, prasanna\}@usc.edu,
rajgopal.kannan.civ@army.mil
}
}

\maketitle
\thispagestyle{arxivfirstpage}

\begingroup
\renewcommand{\thefootnote}{\fnsymbol{footnote}}
\footnotetext[1]{These authors contributed equally.}
\footnotetext[2]{Corresponding author.}
\endgroup

\begin{abstract}
Graph analytics over structured complex networks generates irregular, topology-dependent memory-access streams via indirect traversals, limiting the effectiveness of data prefetching. Accurate prediction enables timely prefetches that hide memory latency. However, existing machine-learning predictors couple short histories with fixed horizons, fitting the training trace’s node layout rather than learning transferable graph patterns. We propose LoCoMAP, a long-context representation-learning framework for memory-access prediction with independently controlled observation history and prediction horizon. We evaluate Connected Components (CC), PageRank (PR), and Single-Source Shortest Paths (SSSP) on six synthetic graph families, with Erdős–Rényi (ER) and Grid as random and deterministic controls. We test two interventions: same-family regeneration changes connectivity, while node shuffling disrupts layout but preserves connectivity. Across 18 graph–workload pairs, extending history from 8 to 128 accesses improves F1 by 7.8\%, while extending prediction horizon from 8 to 512 improves F1 by 88.3\%. For CC, regeneration is less disruptive than shuffling: without retraining, long context retains 95.4\% of original F1 and outperforms short context in six families. Under full shuffle, long context improves CC F1 by 33.6\% across ER, Barabási–Albert, Copying Model, and Stochastic Block Model (SBM), with a 2.06$\times$ gain on SBM, but not on Watts–Strogatz or Grid, where near-zero delta mass shifts into heavy tails beyond the predictor’s range. Long context outperforms short context on six real-world CC graphs. At full shuffle, long-context PR retains 57.6\% of original F1 and averages 11.0\% above short context. Both SSSP models retain about 30\% of original F1 under full shuffle as distance updates access vertex data in a new order. Within these bounds, long-context predictors are robust across related graphs and layouts without retraining, enabling a generalizable graph prefetcher.
\end{abstract}

\begin{IEEEkeywords}
data prefetching, graph analytics, graph topology, memory access prediction, long-context modeling
\end{IEEEkeywords}

\section{Introduction}

Data prefetching hides memory latency by predicting future memory accesses and fetching data into the cache hierarchy before the processor requests it~\cite{dubois2012parallel, vanderwiel2000data, byna2008taxonomy}. Its effectiveness rests on accurate memory access prediction: correct and timely predictions overlap latency with computation, while incorrect ones waste bandwidth and pollute caches~\cite{srinath2007feedback}. Rule-based hardware prefetchers exploit simple regularities such as streams and strides but fail when accesses become irregular~\cite{smith1978sequential, jouppi1990improving, somogyi2006spatial, michaud2016best}. Machine learning (ML) prefetchers instead learn access patterns as sequence models and achieve substantially higher accuracy on irregular workloads~\cite{hashemi2018learning, shi2021a, zhang2022fine}, yet graph analytics remain among the hardest applications to predict~\cite{basak2019analysis, zhang2023phases}.

Graph analytics underpin web search, recommendation systems, and social networks~\cite{drosou2016enhanced, meusel2015graph}, but they execute through indirect memory accesses (e.g., \texttt{row[col[i]]}) whose targets depend on previously loaded values~\cite{lumsdaine2007challenges, ainsworth2016graph}, producing sparse access streams with poor locality that leave graph processing memory-bound~\cite{wulf1995hitting, beamer2017the, basak2019analysis}. This irregularity, however, is not arbitrary~\cite{newman2003structure, albert2002statistical}: real-world networks exhibit well-understood structure, such as power-law degree distributions~\cite{barabasi1999emergence, broder2000graph}, small-world connectivity~\cite{watts1998collective}, and community organization~\cite{girvan2002community}, all of which can be reproduced under controlled conditions by synthetic graph models~\cite{erdos1959on, barabasi1999emergence, kumar2000stochastic, abbe2017community, watts1998collective}. A predictor that captures this structure can use topology as an exploitable prior for memory access prediction.

Existing ML predictors do not exploit this prior. State-of-the-art approaches such as TransFetch~\cite{zhang2022fine}, SHARP~\cite{zhang2022sharp}, and MPGraph~\cite{zhang2023phases} apply a single short-context configuration, with fewer than ten accesses of history, uniformly to all graph workloads, implicitly assuming that all predictive signal lies within a short window and that one configuration fits every graph structure. Neither assumption holds: different topologies induce fundamentally different delta distributions and temporal dependencies. Scale-free graphs~\cite{barabasi1999emergence} revisit hubs at long, irregular intervals; community-structured graphs~\cite{girvan2002community, abbe2017community} produce clustered access bursts that exceed any short window; small-world graphs~\cite{watts1998collective} interleave local traversal with long-range jumps; and spatial grids~\cite{barthelemy2011spatial} are nearly deterministic. A short-context predictor cannot distinguish these regimes and instead memorizes the vertex layout of its training trace.

Hence, we propose \ourwork, a long-context memory access prediction framework for graph analytics. \ourwork~executes Single-Source Shortest Paths, PageRank, and Connected Components from the GAP benchmark suite~\cite{beamer2017the} on six synthetic graph families: Erd\H{o}s--R\'enyi graphs and spatial grids as controls at the extremes of pure randomness and perfect regularity, and Barab\'asi--Albert, Copying Model, stochastic block model, and Watts--Strogatz graphs as representatives of the major structural classes of real-world networks. \ourwork~formulates prediction as a long-context representation learning problem in which the look-back length $L$, the look-forward window $W$, and the delta range $D$ are independent parameters, separating how much history a predictor needs from how far its predictions should reach. Because longer contexts raise inference cost under the strict latency budgets of ML prefetching~\cite{shi2021a, zhang2024attention, gupta2023packd, zhang2026net2tab}, we characterize the latency and FLOPs cost of extending context alongside its accuracy gains. Finally, we test whether the learned signal is genuinely topological by evaluating trained predictors under progressively shuffled node orderings, which destroy memory locality while preserving connectivity~\cite{wei2016speedup, balaji2018graph}.
We summarize our main contributions as follows:

\begin{itemize}

\item We propose \ourwork, a long-context memory-access prediction framework that maps access histories to multi-hot bitmaps of future address deltas. Independently controlling observation history, prediction horizon, and delta range separates the context used to form the representation from the temporal and spatial scope of its predictions.

\item We construct a topology-controlled evaluation of CC, PR, and SSSP across six synthetic graph families. Erdős–Rényi and Grid provide random and deterministic controls; Barab\'asi--Albert, Copying Model, stochastic block model, and Watts--Strogatz capture hub-dominated growth, link copying, community organization, and small-world connectivity across biological, web, social, and infrastructure networks. Same-family regeneration tests unseen connectivity, while progressive node shuffling tests disrupted layouts. Gnutella peer-to-peer, Skitter Internet, Google web, DBLP collaboration, FlyWire connectome, and roadCA networks extend the CC evaluation to real graphs.

\item We quantify the accuracy--cost tradeoff of long context. At fixed prediction horizon, increasing history from 8 to 128 accesses improves geometric-mean F1 by 7.8\% across 18 graph--workload pairs and by up to 36.8\%, while requiring $21\times$ the per-inference FLOPs and $6.3\times$ the latency. At fixed history, increasing the prediction horizon from 8 to 512 accesses improves geometric-mean F1 by 88.3\%.

\item We identify the limits of predictor reuse without retraining. Long-context CC outperforms short-context on all held-out same-family instances and six fully shuffled real graphs, improving mean F1 on the latter by 10.2\%. Under full shuffle, it improves geometric-mean F1 by 33.6\% on ER, BA, CM, and SBM and reaches $2.06\times$ short-context F1 on SBM, but loses its advantage on WS and Grid, where delta mass moves beyond the output range. PR retains a long-context advantage on average, but SSSP does not.

\end{itemize}

\section{Background}
\subsection{Graph Analytics and Synthetic Graph Topologies}

Graph analytics computes structural properties, such as shortest-path distances, vertex importance, and connectivity, by traversing a graph according to its connectivity~\cite{page1999pagerank, malewicz2010pregel}. Because traversal order is determined by topology rather than memory layout, the resulting access streams are indirect, with locality that varies across applications and graph structures~\cite{lumsdaine2007challenges, beamer2017the, basak2019analysis}.
  \begin{figure}[t!]
    \centering
    \includegraphics[width=\linewidth]{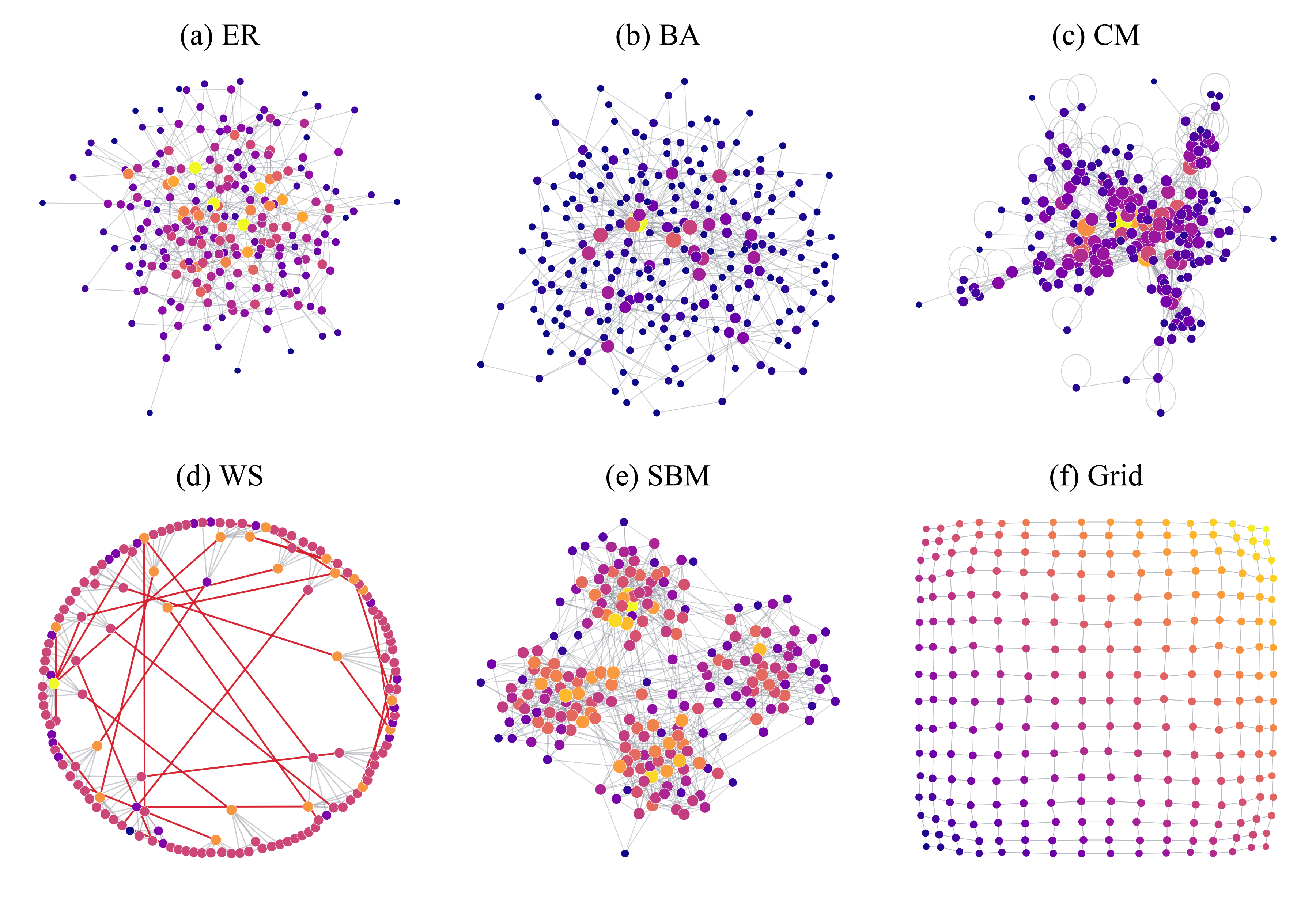}
    \caption{Graph Topologies of Synthetic Network Models.}
    \label{fig:topologies} 
\end{figure}
Synthetic graph models isolate the influence of topology on memory behavior while avoiding the confounding factors of real-world datasets~\cite{chakrabarti2004rmat, newman2003structure}. As illustrated in Fig.~\ref{fig:topologies}, we consider six families, ordered from independent random connectivity to deterministic spatial regularity:

\noindent\textbf{Erd\H{o}s--R\'enyi (ER):} Erd\H{o}s--R\'enyi graphs serve as baseline models for random communication networks and randomized distributed systems~\cite{byna2008taxonomy}. The $G(n,p)$ model includes each possible edge independently with probability $p$~\cite{erdos1959on}, yielding a binomial degree distribution:
\begin{equation}
P(k) = \binom{n-1}{k} p^{k} (1-p)^{n-1-k}
\label{eq:er}
\end{equation}
where $n$ is the number of vertices and $p$ is the edge inclusion probability. The result is homogeneous random connectivity with no hubs and no community structure.
 
\noindent\textbf{Barab\'asi--Albert (BA):} Barab\'asi--Albert graphs model scale-free networks with hub-dominated connectivity, including protein interaction networks, citation networks, and Internet autonomous systems~\cite{meusel2015graph}. New vertices attach to existing vertices with probability proportional to their degree~\cite{barabasi1999emergence}:
\begin{equation}
\Pi(k_i) = \frac{k_i}{\sum_{j} k_j}
\label{eq:ba}
\end{equation}
where $\Pi(k_i)$ is the probability that a new vertex connects to vertex $i$ of current degree $k_i$. Preferential attachment yields a power-law degree distribution $P(k) \propto k^{-3}$ dominated by a small number of high-degree hubs~\cite{barabasi1999emergence}.
 
\noindent\textbf{Copying Model (CM):} Copying Model graphs capture the hyperlink structure of the World Wide Web and related information networks, including citation graphs and online document collections~\cite{kumar2000stochastic}. Each arriving vertex selects a random prototype $u$ and forms its $i$-th out-link either uniformly at random or by copying the prototype's $i$-th link:
\begin{equation}
    \Pr(v_i=w \mid u_i)
    =
    \frac{\alpha}{n}
    +
    (1-\alpha)\mathbf{1}\{w=u_i\}.
    \label{eq:copying_model}
\end{equation}
where $v_i$ is the destination of the new vertex's $i$-th link, $u_i$ is the $i$-th neighbor of the prototype $u$, $n$ is the current number of vertices, and {$\alpha$ is the probability of selecting a destination uniformly at random, while $1-\alpha$ is the probability of copying $u_i$. The indicator function $\mathbf{1}\{w=u_i\}$ equals one when $w=u_i$ and zero otherwise.} Copying yields power-law degrees and locally dense link patterns that emerge stochastically rather than from an explicit global partition~\cite{kumar2000stochastic}.

\noindent\textbf{Stochastic Block Model (SBM):} Stochastic Block Model graphs model community-structured networks such as online social communities, scientific collaboration networks, and protein interaction networks~\cite{girvan2002community}.  Vertices are partitioned into $K$ communities by an assignment $z: V \rightarrow \{1,\dots,K\}$, and each edge is included independently with probability determined by the community memberships of its endpoints:
\begin{equation}
\Pr\left[(u,v) \in E\right] = B_{z(u)z(v)}
\label{eq:sbm}
\end{equation}
where $\Pr\left[(u,v) \in E\right]$ is the probability of an edge between vertices $u$ and $v$, and $B_{z(u)z(v)}$ is the block connection probability for their communities $z(u)$ and $z(v)$. Setting $B_{ii} > B_{ij}$ for $i \neq j$ produces densely connected clusters joined by sparse inter-community bridges~\cite{holland1983stochastic, abbe2017community}. 
 
\noindent\textbf{Watts--Strogatz (WS):} Watts--Strogatz graphs model small-world networks, including power grids, neuronal networks, and social networks~\cite{watts1998collective}. Starting from a ring lattice, edges are rewired with probability $p$; for intermediate $p$, the graph combines short paths of a random graph with local clustering:
\begin{equation}
L(p) \gtrsim L_{\mathrm{random}}, \qquad C(p) \gg C_{\mathrm{random}}
\label{eq:ws}
\end{equation}
where $L(p)$ and $C(p)$ denote the path length and clustering coefficient at rewiring probability $p$, and $L_{\mathrm{random}}$ and $C_{\mathrm{random}}$ the corresponding quantities for an equivalent random graph.

\noindent\textbf{Spatial Grid:} Grid graphs represent spatially structured systems such as image pixels, sensor arrays, road networks, and integrated circuit layouts~\cite{basak2019analysis}. Vertices lie on a two-dimensional lattice with coordinates $(x_u, y_u)$ and are adjacent exactly at unit Manhattan distance:
\begin{equation}
(u,v) \in E \iff |x_u - x_v| + |y_u - y_v| = 1
\label{eq:grid}
\end{equation}
where $(u,v) \in E$ denotes an edge between vertices $u$ and $v$. The result has uniform degree and strong spatial locality~\cite{barthelemy2011spatial}. 
 
Together, these families span the structural spectrum relevant to our evaluation, from independent randomness to increasingly explicit local and mesoscopic organization~\cite{newman2003structure, albert2002statistical}.

\subsection{Memory Access Prediction}
Memory access prediction estimates future memory references from execution history~\cite{hashemi2018learning, zhang2022sharp}. Following recent ML-based prefetching work~\cite{shi2021a, zhang2022fine}, we represent each access by its program counter (PC), memory block address, and address delta relative to previous accesses: the PC identifies the referencing instruction, block addresses reduce granularity to cache-line resolution, and deltas encode relative movement that exposes recurring patterns less apparent in absolute addresses~\cite{hashemi2018learning, zhang2024attention}. Together, these features provide  spatial and program-semantic cues for learning irregular access behavior.

Graph-processing frameworks improve locality through graph partitioning, vertex reordering, and cache-aware data layouts~\cite{gonzalez2012powergraph,shun2013ligra,wei2016speedup,
kyrola2012graphchi,balaji2018graph,beamer2015locality,
mukkara2018exploiting}, but largely ignore the access-pattern differences induced by graph topology. 
We instead use topology-controlled graph families to study whether long context captures topology-induced regularities.

\section{Related Work}
\subsection{Rule-Based and ML-Based Data Prefetching}
Data prefetching hides memory latency by predicting future memory accesses and bringing data into the cache hierarchy before it is requested by the processor~\cite{smith1978sequential, vanderwiel2000data, byna2008taxonomy}. Effective prefetching therefore depends on accurate memory access prediction~\cite{srinath2007feedback}. For graph analytics, sparse graph structures and indirect memory references generate irregular access patterns that significantly reduce the effectiveness of conventional hardware prefetchers~\cite{ainsworth2016graph, ainsworth2018software, basak2019analysis}.
  \begin{figure*}[!t]
    \centering
    \includegraphics[width=0.7\linewidth]{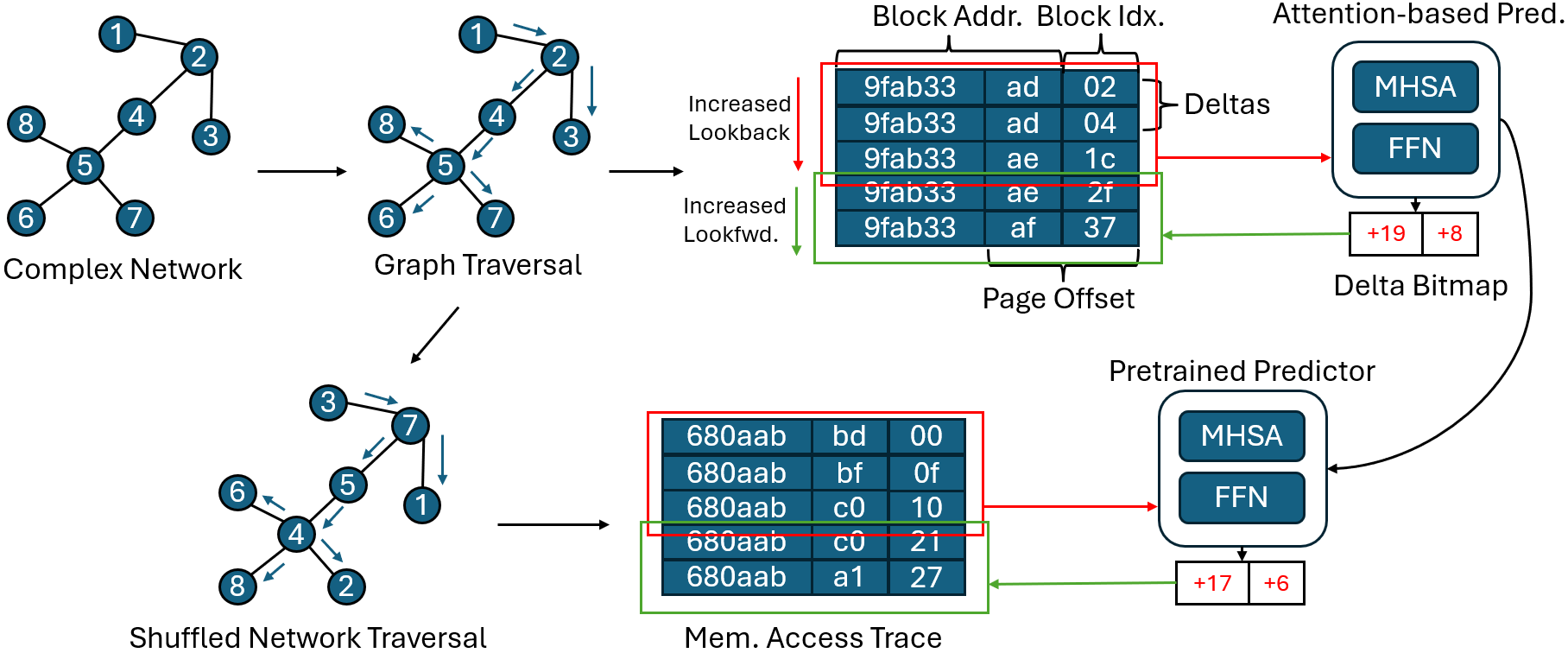}
    \caption{Overall Architecture of \ourwork~.}
    \label{fig:overall}
\end{figure*}
\subsubsection{Rule-Based Prefetchers}
Existing data prefetching approaches can be broadly divided into rule-based and machine learning (ML)-based methods. Rule-based prefetchers, including stride, stream, and Best-Offset (BO)~\cite{chen1995effective, jouppi1990improving, michaud2016best}, exploit spatial or temporal locality using predefined heuristics~\cite{somogyi2006spatial, jain2013linearizing} and achieve good performance on regular workloads. However, graph analytics generate sparse and indirect memory accesses that exhibit little predictable locality, limiting the effectiveness of these rule-based designs~\cite{beamer2017the, basak2019analysis}. To address this challenge, graph-aware prefetchers such as the Indirect Memory Prefetcher (IMP) exploit pointer-based memory references to improve prediction accuracy for graph workloads~\cite{yu2017imp,zhang2025grafetch}. However, IMP is primarily designed for L1 cache prefetching and does not address memory access prediction for last-level cache (LLC) prefetching, where accurate prediction matters most: LLC prefetching hides DRAM accesses costing hundreds of cycles, whereas L1/L2 prefetching hides comparatively fast L2/LLC accesses. Consequently, rule-based approaches remain limited in capturing the complex long-context memory access behavior of irregular graph analytics.

\subsubsection{ML-Based Prefetchers}
Recently, ML-based prefetchers formulate memory access prediction as a sequence modeling problem~\cite{hashemi2018learning}. Representative approaches include Voyager~\cite{shi2021a}, TransFetch~\cite{zhang2022fine}, and MPGraph~\cite{zhang2023phases}, which learn memory access patterns directly from execution traces instead of relying on fixed heuristics. These methods have demonstrated significantly higher prediction accuracy than traditional prefetchers on irregular workloads~\cite{zhang2022sharp, zhang2023phases}. However, their inference latency and storage requirements exceed the strict timeliness constraints of hardware prefetching, especially at the LLC; recent efforts such as Net2Tab narrow this gap by distilling attention-based predictors into compact hierarchies of tables~\cite{zhang2024attention, zhang2026net2tab}.
 
Existing ML-based approaches, however, fix one short-context model for all workloads and do not examine how topology shapes access behavior. In
this work, we treat execution history as a controllable parameter,
evaluated across topology-controlled graph families,
and use node shuffling to separate learned structure from memorized
layout.

\section{Approach}

\ourwork~formulates memory access prediction as a long-context sequence-learning problem and evaluates how much information longer context provides across graph topologies.

\subsection{Problem Formulation.}
Let $X_t = \{x_1, x_2, \dots, x_L\}$ be the sequence of $L$ history block
addresses at time $t$, where $x_i = a_i \gg \beta$ discards the $\beta$-bit
block offset from the raw address $a_i$. Let $P_t = \{\mathrm{pc}_1,
\mathrm{pc}_2, \dots, \mathrm{pc}_L\}$ be the sequence of $L$ history program
counters at time $t$, and let $I_t = \{(x_1,\mathrm{pc}_1), \dots,
(x_L,\mathrm{pc}_L)\}$ be the input of the prediction model. Let $O_t =
\{o_1, o_2, \dots, o_{2D}\}$ be the multi-hot label over the future block
deltas within range $[-D,+D]$ occurring across the next $W$ accesses. Our
goal is to learn a meaningful mapping from $I_t$ to $O_t$ for data
prefetching; final address predictions are the addition of current
block address and predicted deltas. Unlike prior work that fixes a short
history, we treat look-back length $L$ as a long context to expose long-range dependencies in the access stream.

\subsection{Overall Design}

Fig.~\ref{fig:overall} illustrates the overall workflow of \ourwork. A synthetic graph is executed using a GAP workload to generate an LLC memory-access trace. Each memory access is converted to a cache block address and represented using fine-grained address segmentation and execution context. An attention-based predictor takes the previous $L$ accesses as input and predicts a bitmap of future address deltas within a look-forward window $W$ and delta bound $D$. The predicted deltas are then added to the current block address to recover multiple future memory accesses. 
The $L$ history block addresses are processed by fine grained address segmentation, and
the program counters are hashed and embedded as context. The
segmented addresses are mapped by a linear projection, combined with
position and context embeddings, and encoded by Transformer
layers~\cite{vaswani2017attention}, whose self attention relates
every access to the entire $L$-access history in a single step. A
classification head with sigmoid activation outputs the probability
$p_i$ of each of the $2D$ deltas, trained with binary cross entropy
over the independent labels:
\begin{equation}
\mathcal{L} = -\frac{1}{2D}\sum_{i=1}^{2D}
\big[\, o_i \log p_i + (1-o_i)\log(1-p_i) \,\big].
\label{eq:loss}
\end{equation}
At inference, the output probabilities are binarized with a
validation selected threshold that maximizes F1-score, each
predicted delta yields the prefetch address $x_L$ plus that delta,
so a single inference produces multiple prefetch candidates.

\begin{figure}[H]
    \centering
    \includegraphics[width=1.00\linewidth]{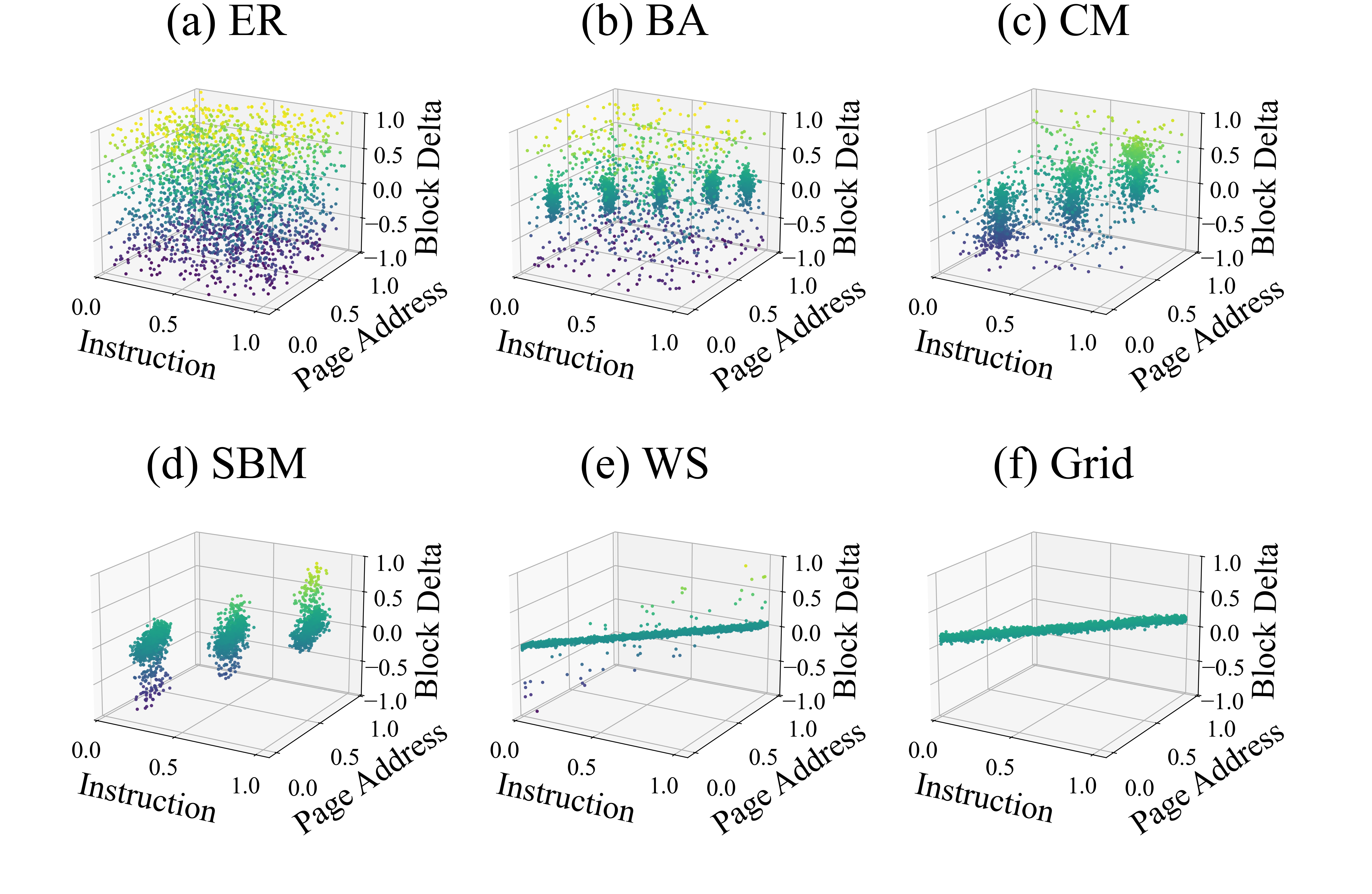}
    \caption{Visualization of memory access patterns in the benchmark applications.}
    \label{fig:traces}
\end{figure}

\section{Evaluation}
\subsection{Experimental Setup}
\subsubsection{Datasets}

{
We evaluate \ourwork~using six synthetic graph families
that capture structural properties observed in real-world
complex networks as well as six real-world graphs as summarized in Table~\ref{tab:graph_stats}.
}
The graphs are represented in compressed sparse row (CSR) format for
algorithm processing.

\begin{table}[htbp]
\centering
\caption{Graph statistics.}
\label{tab:graph_stats}
\begin{tabular}{|l|r|r|c|l|}
\hline
\textbf{Topology} & \textbf{\# Vertices} & \textbf{\# Edges} &
\textbf{Avg. Degree} & $\mathbf{S}_{\mathrm{order}}$ \\
\hline
ER   & 1,000,000& 2,400,000 & 4.8  &0.0007\\
\hline
BA   & 1,000,000 & 4,999,985 & 10.0  &0.0016\\
\hline
CM   & 1,000,000 & 2,952,366 & 5.9  &0.2088\\
\hline
SBM  & 1,000,000   & 2,500,000 & 5.0  &0.5188\\
\hline
WS   & 1,000,000 & 2,000,000 & 4.0  &0.9002\\
\hline
Grid & 1,000,000 & 1,998,000 & 4.0  &0.9985\\
\hline
FlyWire & 134,013 & 2,313,472 & 34.53  &0.0083\\
\hline
Google & 875,713 & 4,322,051 & 9.87  &0.0100\\
\hline
DBLP & 317,080 & 1,049,866 & 6.62  &0.3632\\
\hline
Gnutella & 62,586 & 147,892 & 4.73  &0.4644\\
\hline
Skitter & 1,696,415 & 11,095,298 & 13.08  &0.4730\\
\hline
roadCA & 1,965,206 & 2,766,607 & 2.82  &0.9937\\
\hline
\end{tabular}
\end{table}

\subsubsection{Algorithms}
We implemented three widely used graph processing algorithms based on GPOP framework~\cite{lakhotia2020gpop}.

\begin{itemize}
    \item Single-Source Shortest Paths (SSSP) computes the shortest-path distance from a source vertex to all other vertices by propagating tentative distances along edges until convergence~\cite{dijkstra1959note}.
    \item PageRank (PR) iteratively computes a vertex-importance score by distributing rank mass from each vertex to its neighbors until the scores converge~\cite{page1999pagerank}.
   \item Connected Components (CC) labels each vertex by its component using label propagation, where each vertex adopts the minimum label among its neighbors~\cite{shiloach1982an}.
\end{itemize}

\subsection{Metrics}
We report $\mathrm{F1}=2PR/(P+R)$, the harmonic mean of precision $P$
and recall $R$~\cite{powers2020evaluation}.

Unless stated otherwise, aggregate improvements are geometric means of
per-topology--workload F1 ratios, and ``up to'' refers to the maximum
single-pair ratio.

To quantify the organization of the native vertex ordering, we first
compute the normalized mean edge span
\begin{equation}
\bar{s}
=
\frac{1}{|E|}
\sum_{(u,v)\in E}
\frac{|u-v|}{|V|-1}.
\label{eq:mean_edge_span}
\end{equation}
We then compare the native edge span with its randomly relabeled value and
compute the ordering score as
\begin{equation}
\begin{aligned}
R_{\mathrm{edge}}
&=
\frac{\bar{s}_{\mathrm{native}}}
     {\bar{s}_{\mathrm{random}}}, \\[2pt]
S_{\mathrm{order}}
&=
\frac{\left|R_{\mathrm{edge}}-1\right|}
     {\max\!\left(R_{\mathrm{edge}},1\right)}.
\end{aligned}
\label{eq:ordering_score}
\end{equation}
Here, $\bar{s}_{\mathrm{native}}$ and $\bar{s}_{\mathrm{random}}$ denote
the normalized mean edge spans before and after complete random relabeling.
An $S_{\mathrm{order}}$ near zero indicates a random-like 
assignment, whereas a value closer to one indicates organization
in the CSR layout.

\subsection{Controlled Graph and Trace Generation}
We generate all graphs synthetically using the standard
generative model of each family, with a fixed target size and
comparable average degree, so families differ in structure rather
than scale; each graph is stored in CSR format for processing.

To obtain memory-access traces, we develop a Pintool
using Intel Pin~\cite{luk2005pin} that detects the memory-access operations
executed by each graph workload and records the corresponding
context information, including the accessed virtual address and
program counter. We then replay the collected accesses through
a four-core ChampSim~\cite{champsim} model with a shared last-level cache
(LLC) and record the resulting per-core LLC access stream,
which forms the input to memory-access prediction.

Across the 18 graph-workload traces, we observe 14.1K-5.19M LLC accesses, 276-131.0k unique block addresses, 600-4.1K pages, and 1.8k-613.5K unique deltas.
\begin{table}[!t]
\centering
\caption{Model configuration.}
\label{tab:model_config}
\setlength{\tabcolsep}{5pt}
\begin{tabular}{lllll}
\hline
 & \textbf{Configuration} & \textbf{Value} & \textbf{Configuration} & \textbf{Value} \\
\hline
Input/output & Delta bound   & $\pm$128 & Delta bitmap $B$ & 256 \\
             & History $N$   & 9        & Look-forward $W$ & 256 \\
\hline
Attention    & Dimension $D$ & 64& MLP head layer   & 1   \\
             & Head number   & 4        & Layer $L$        & 1\\
\hline
\end{tabular}
\end{table}
 \subsection{Long-Context Configuration and Cost}
\label{sec:predictor}
\begin{table*}[!htbp]
\centering
\caption{Ablation of Look-Back Length $L$ and Look-Forward Window $W$ Across Graph Topologies and Applications (F1-Score)}
\label{tab:ablation_lw}
\setlength{\tabcolsep}{3pt}
\begin{tabular}{|c|c|c|c|c|c|c|c|c|c|c|c|c|c|c|c|c|c|c|c|}
\hline
\multirow{3}{*}{\textbf{Parameter}} & \multirow{3}{*}{\textbf{Value}} & \multicolumn{18}{c|}{\textbf{Applications}} \\
\cline{3-20}
 & & \multicolumn{6}{c|}{\textbf{CC}} & \multicolumn{6}{c|}{\textbf{PR}} & \multicolumn{6}{c|}{\textbf{SSSP}} \\
\cline{3-20}
 & & \textbf{ER} & \textbf{BA} & \textbf{CM} & \textbf{SBM} & \textbf{WS} & \textbf{Grid}
   & \textbf{ER} & \textbf{BA} & \textbf{CM} & \textbf{SBM} & \textbf{WS} & \textbf{Grid}
   & \textbf{ER} & \textbf{BA} & \textbf{CM} & \textbf{SBM} & \textbf{WS} & \textbf{Grid} \\
\hline
\multirow{5}{*}{\shortstack{Look-back $L$\\($W{=}256$, $B{=}256$)}}
 & 8   & 0.466 & 0.561 & 0.660 & 0.590 & 0.690 & 0.641& 0.472 & 0.658 & 0.684 & 0.609 & 0.733 & 0.668& 0.075 & 0.256 & 0.683 & 0.057 & 0.363 & 0.639\\
\cline{2-20}
 & 16  & 0.468 & 0.574 & 0.661 & 0.592 & 0.691 & 0.642& 0.475 & 0.669 & 0.688 & 0.613 & 0.738 & 0.671& 0.083 & 0.273 & 0.684 & 0.065 & 0.354 & 0.646\\
\cline{2-20}
 & 32  & 0.503 & 0.568 & 0.664 & 0.593 & 0.714 & 0.645& 0.492 & 0.661 & 0.690 & 0.616 & 0.740 & 0.665& 0.086 & 0.303 & 0.696 & 0.066 & 0.365 & 0.653\\
\cline{2-20}
 & 64  & \textbf{0.515} & 0.620 & 0.669 & \textbf{0.596} & \textbf{0.731} & \textbf{0.650}& 0.503 & 0.670 & 0.691 & \textbf{0.622} & 0.748 & 0.679& 0.093 & 0.311 & 0.704 & 0.074 & 0.358 & 0.655\\
\cline{2-20}
 & 128 & 0.514 & \textbf{0.635} & \textbf{0.674} & 0.587 & 0.722 & 0.648& \textbf{0.520} & \textbf{0.682} & \textbf{0.693} & 0.619 & \textbf{0.757} & \textbf{0.681}& \textbf{0.094} & \textbf{0.312} & \textbf{0.716} & \textbf{0.078} & \textbf{0.369} & \textbf{0.663}\\
\hline
\multirow{5}{*}{\shortstack{Look-forward $W$\\($L{=}8$, $B{=}256$)}}
 & 8   & 0.375 & 0.209 & 0.369 & 0.255 & 0.150 & 0.263& 0.355 & 0.435 & 0.372 & 0.394 & 0.325 & 0.373& 0.049 & 0.235 & 0.594 & 0.066 & 0.170 & 0.537\\
\cline{2-20}
 & 32  & 0.361 & 0.391 & 0.486 & 0.428 & 0.350 & 0.402& 0.389 & 0.572 & 0.464 & 0.513 & 0.551 & 0.527& 0.065 & 0.243 & 0.600 & 0.057 & 0.339 & 0.621\\
\cline{2-20}
 & 128 & 0.431 & 0.514 & 0.589 & 0.549 & 0.607 & 0.581& 0.432 & 0.619 & 0.601 & 0.582 & 0.694 & 0.605& 0.092 & 0.251 & 0.648 & 0.075 & 0.369 & 0.625\\
\cline{2-20}
 & 256 & 0.466 & 0.561 & 0.660 & 0.590 & 0.690 & 0.641& 0.472 & 0.658 & 0.684 & 0.609 & 0.733 & 0.668& 0.075 & 0.256 & \textbf{0.683} & 0.057 & 0.363 & 0.639\\
\cline{2-20}
 & 512 & \textbf{0.507} & \textbf{0.606} & \textbf{0.710} & \textbf{0.623} & \textbf{0.735} & \textbf{0.696}& \textbf{0.501} & \textbf{0.693} & \textbf{0.736} & \textbf{0.620} & \textbf{0.740} & \textbf{0.713}& \textbf{0.095} & \textbf{0.259} & 0.682 & \textbf{0.118} & \textbf{0.388} & \textbf{0.654}\\
\hline
\end{tabular}
\end{table*}
We investigate long-context memory access prediction by varying the
look-back length $L$ and the look-forward window $W$. Increasing the look-back length $L$ lengthens the input sequence, regularities at any temporal distance within the window, such as hub revisitation
and community dwell intervals, become directly reachable by
attention. Increasing the look-forward window $W$ aggregates deltas over a
longer future horizon, so the bitmap becomes denser and long-range
predictive information is converted into additional coverage.

\begin{figure}[H]
    \centering
    \includegraphics[width=0.95\linewidth]{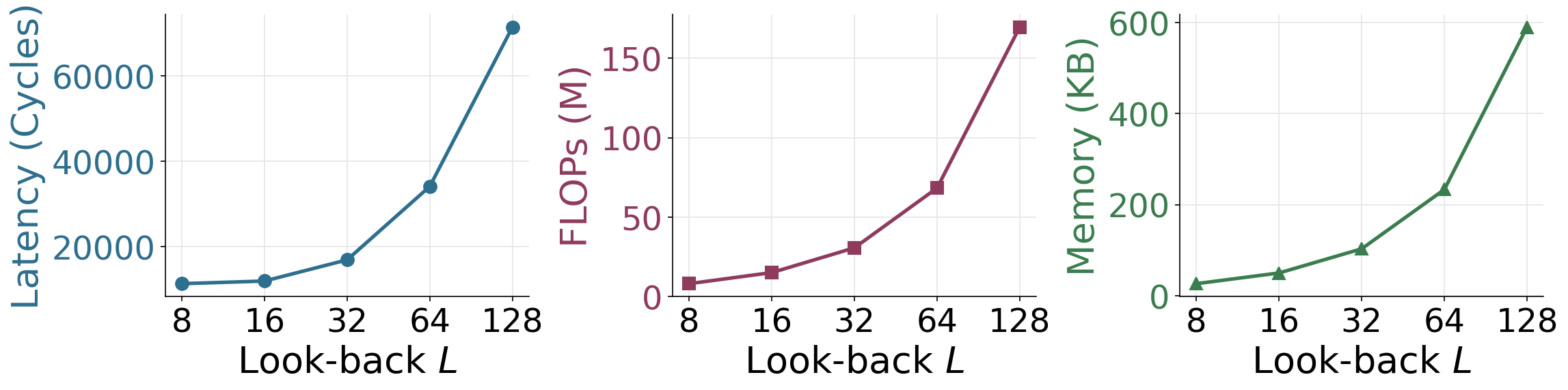}
    \caption{Latency, FLOPs and memory cost as the look-back length $L$ scales from 8 to 128.}
    \label{fig:cost}
\end{figure}

However, the representation gains of long context are not free.
Fig.~\ref{fig:cost} reports inference FLOPs and latency as the
look-back length scales from $L{=}8$ to $L{=}128$. FLOPs per
inference grow from {8M} to
{169.6M}, a {21$\times$}
increase, and latency grows from {11{,}357} to
{71{,}345} {[cycles]}, a
{6.3$\times$} increase, reflecting the quadratic
cost of self attention at large $L$. Setting these costs against
Table~\ref{tab:ablation_lw} yields the practical tradeoff: for
several topology and workload pairs the accuracy gains largely
arrive by $L{=}64$, while cost continues to grow quadratically
beyond it, so intermediate context lengths capture much of the
benefit at a fraction of the cost. Established deployment
techniques for ML prefetchers, including model compression and
distance prefetching~\cite{gupta2023packd,gupta2024tabconv,zhang2023phases,
zhang2024attention,zhang2026net2tab}, apply directly to our models and can offset the
remaining latency.

 \subsection{Training and Evaluation}
For each graph--workload pair, we split the collected memory trace into training and testing sets and train a delta predictor to map the access history to a bitmap of future address deltas.

\section{Results}

\subsection{Ablation Study}
We conduct ablation studies on the two long-context parameters, the look-back length $L$ and the look-forward window $W$, and report the results in Table~\ref{tab:ablation_lw}.

\subsubsection{Effectiveness of look-back Length}
Increasing $L$ from $8$ to $128$ consistently improves prediction accuracy across all workloads. Averaged over the six topologies, the F1-score improves by $5.0\%$ for CC, $3.6\%$ for PR, and $15.0\%$ for SSSP. This shows that additional execution history exposes topology-induced regularities that are not visible within a short context window, motivating long-context prediction for graph analytics.

\subsubsection{Effectiveness of look-forward Window}
The look-forward window is the dominant factor: extending $W$ from $8$ to $512$ improves the F1-score by $2.49\times$ for CC, $76.7\%$ for PR, and $51.7\%$ for SSSP. Because a larger output range converts predictive information into additional coverage, this result confirms that much of the exploitable structure in graph access streams lies beyond a short prediction horizon.

\subsection{{Cross-Instance Generalization}}
{
To evaluate if learned access patterns transfer beyond a single
graph instance, we apply each trained CC predictor to a held-out graph
from the same family; Table~\ref{tab:cross_instance} reports the results.
The long-context model generalizes without retraining to graphs with
unseen connectivity, retaining an average of 94.9\% of its
original-graph F1-score across the five stochastic families. Across all
six families, it achieves higher held-out F1-score and raises mean
retention from 92.5\% to 95.4\%. Retention improves for five families,
with BA as the sole exception.
}

\begin{table}[H]
\centering
\caption{Cross-instance CC generalization without retraining. Each entry
reports Instance A F1/held-out same-family Instance B F1}
\label{tab:cross_instance}

\footnotesize
\renewcommand{\arraystretch}{1.08}

\begin{tabular*}{0.85\columnwidth}
{@{\extracolsep{\fill}}lcc@{}}
\hline
& Original ($L=8$) & Long ($L=128$) \\
Family & A/B (ret.) & A/B (ret.) \\
\hline
ER   & .466/.393 (84.3\%) & .514/.481 (93.6\%) \\
BA   & .561/.546 (97.3\%) & .635/.589 (92.8\%) \\
CM   & .660/.560 (84.8\%) & .674/.622 (92.3\%) \\
SBM  & .590/.569 (96.4\%) & .587/.576 (98.1\%) \\
WS   & .690/.656 (95.1\%) & .722/.705 (97.6\%) \\
Grid & .641/.622 (97.0\%) & .648/.634 (97.8\%) \\
\hline
Mean & .601/.558 (92.5\%) & .630/.601 (95.4\%) \\
\hline
\end{tabular*}
\end{table}
\subsection{{Generalization Under Node Shuffling}}
To test whether long context enables the predictor to learn graph
topology rather than a particular vertex ordering, we evaluate each
trained predictor, without retraining, on CC traces generated after
permuting a controlled fraction of vertex identifiers, comparing a
short context model ($L{=}8$) against a long context model
($L{=}128$); results are shown in
Table~\ref{tab:full_shuffle_retention}.

Fig.~\ref{fig:cc_rod} shows that shuffling reshapes each family's
delta distribution in a different way: CM and SBM are
nearly unchanged; ER sharpens toward small deltas; BA lowers its
peak while staying in the same band; and for WS and Grid, the narrow
peak of layout induced locality collapses. Because the changes in distribution span from invariant, centralized to decentralized, long context is needed since our delta ranges are wide
and differ across families. A short window of eight accesses
exposes too few deltas to characterize a family's range, whereas a
long history samples the distribution broadly enough.

Under complete ($100\%$) shuffling, the long context model retains higher accuracy on all
four families representing real-world network classes, with the
largest gap on community-structured SBM graphs, and ER remains the
most robust overall, consistent with its homogeneous random
connectivity. These are exactly the families whose delta
distributions survive shuffling in Fig.~\ref{fig:cc_rod}:
structure that persists after the layout is destroyed is carried by
connectivity, and long context gives the model enough history to
exploit it. In contrast, for WS and Grid, whose accuracy derives
from layout-induced neighborhood locality, both models collapse and
long context confers no advantage. The robustness benefit of long
context is therefore specifically a benefit of learning topology structure.

\begin{table}[!htbp]
\centering
\caption{F1-Score Retained Under Full Node Shuffling}
\label{tab:full_shuffle_retention}
\setlength{\tabcolsep}{6pt}
\begin{tabular}{|l|c|c|}
\hline
\textbf{Graph} &
\textbf{Original Model} &
\textbf{Long Context Model} \\
\hline
ER   & 92.5\% & 93.8\% \\
\hline
BA   & 62.4\% & 63.9\% \\
\hline
CM   & 57.7\% & 67.5\% \\
\hline
SBM  & 34.2\% & 70.7\% \\
\hline
WS   & 31.6\% & 25.6\% \\
\hline
Grid & 40.9\% & 24.5\% \\
\hline
\end{tabular}
\end{table}

\subsection{{When Long Context Improves Robustness}}
\label{sec}
\begin{figure}
    \centering
    \includegraphics[width=1\linewidth]{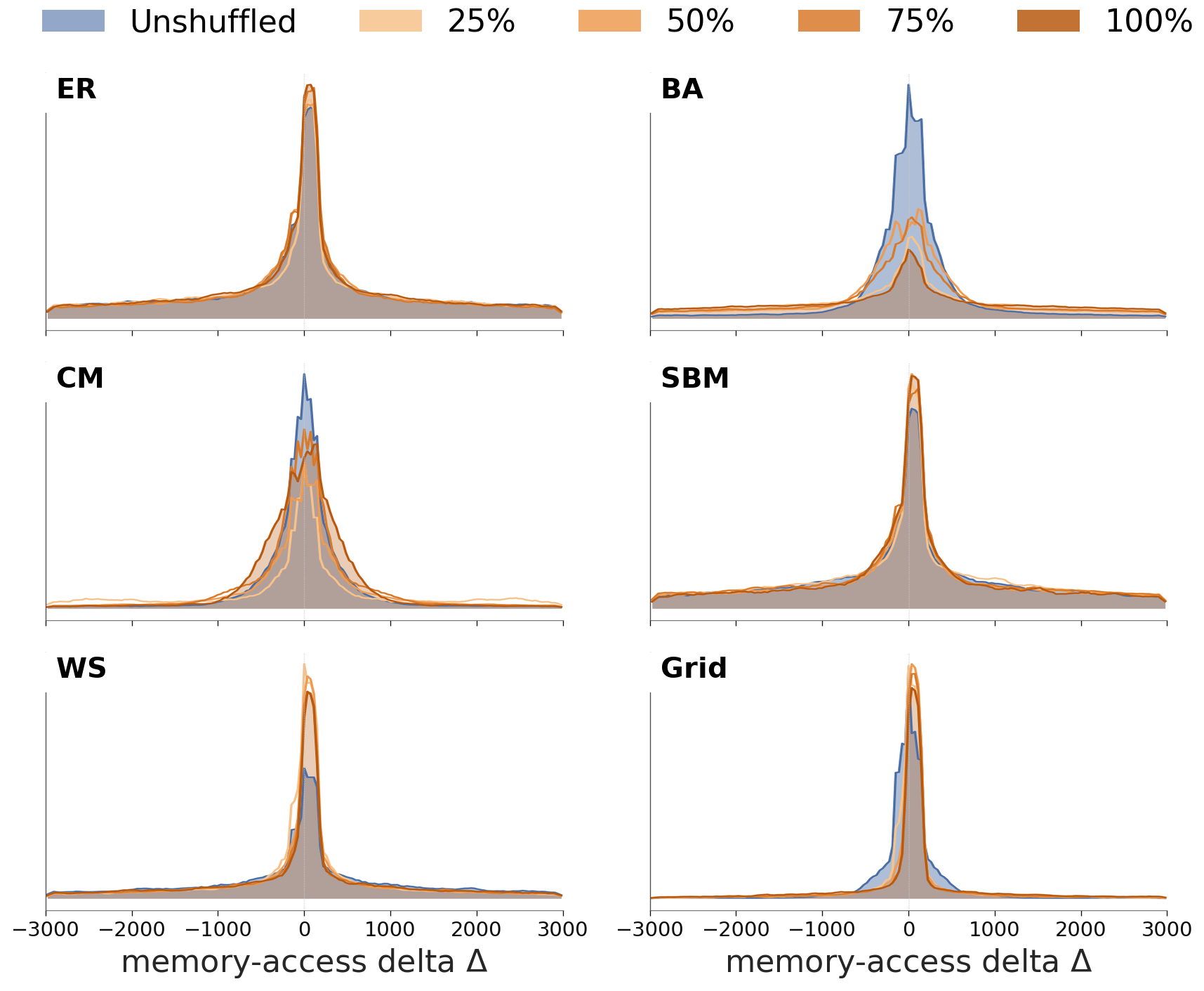}
    \caption{Memory-access delta distributions per graph family, before (blue) and after
(orange) node shuffling for Single-Source Shortest Paths (SSSP) workload.}
    \label{fig:sssp_rod}
\end{figure}

\begin{figure}
    \centering
    \includegraphics[width=.8\linewidth]{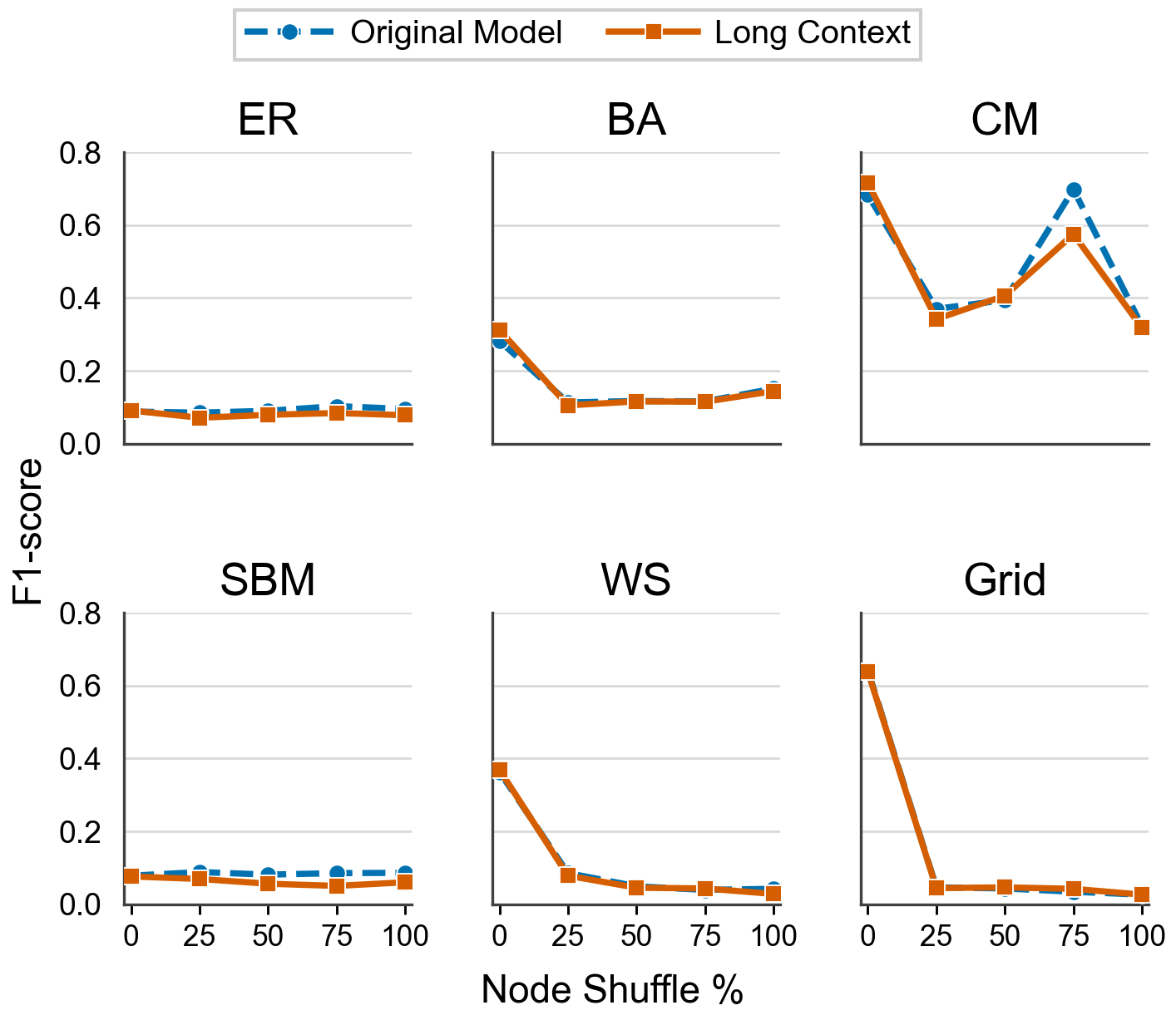}
    \caption{F1-score under node shuffling (L=8 vs. L=128, SSSP
workload).}
    \label{fig:sssp_f1}
\end{figure}

{
Long-context robustness under node shuffling depends on how each
workload's memory accesses change. CC robustness requires deltas to stay within the
predictor's output range. PR remains predictable as shuffling
preserves its full-graph sweep, whereas SSSP loses its data-dependent
visited-set access order. Figs.~\ref{fig:cc_rod}--\ref{fig:sssp_f1}
compare delta distributions and F1 retention.

\noindent\textbf{Connected Components:}
CC propagates minimum component labels across adjacent vertices. Relabeling preserves connectivity but remaps vertex and adjacency addresses, so retention tracks whether deltas remain within the predictor's output range. Long context achieves higher retention than short context on ER, BA, CM, and SBM, with ER retaining 93.8\% of its F1-score. On WS and Grid, near-zero delta mass shifts into heavier tails, reducing long-context retention to 25.6\% and 24.5\%.

\begin{figure}
    \centering
    \includegraphics[width=1\linewidth]{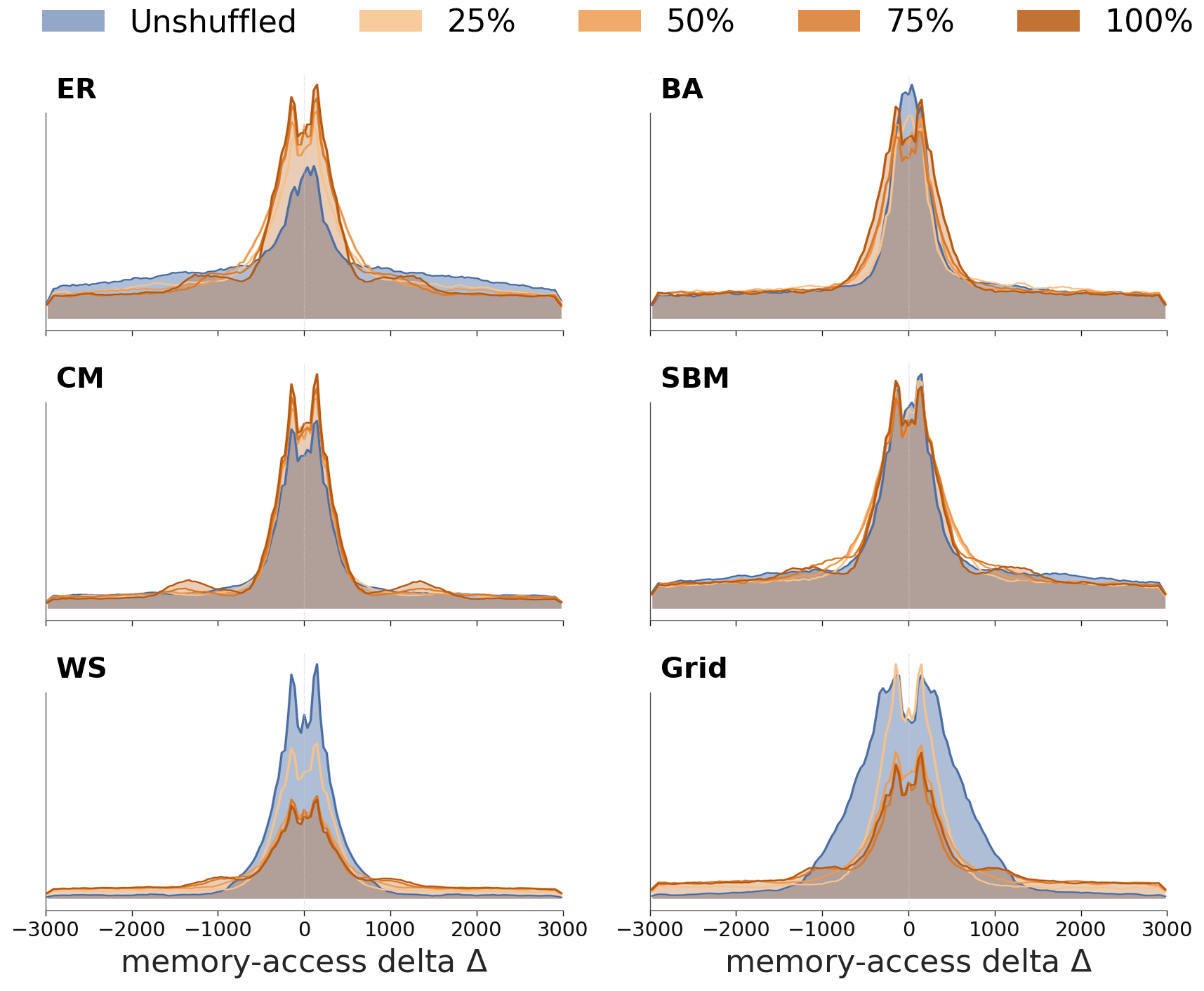}
    \caption{Memory-access delta distributions per graph family, before (blue) and after
(orange) node shuffling for Connected Components (CC) workload.}
    \label{fig:cc_rod}
\end{figure}
\begin{figure}
    \centering
    \includegraphics[width=.8\linewidth]{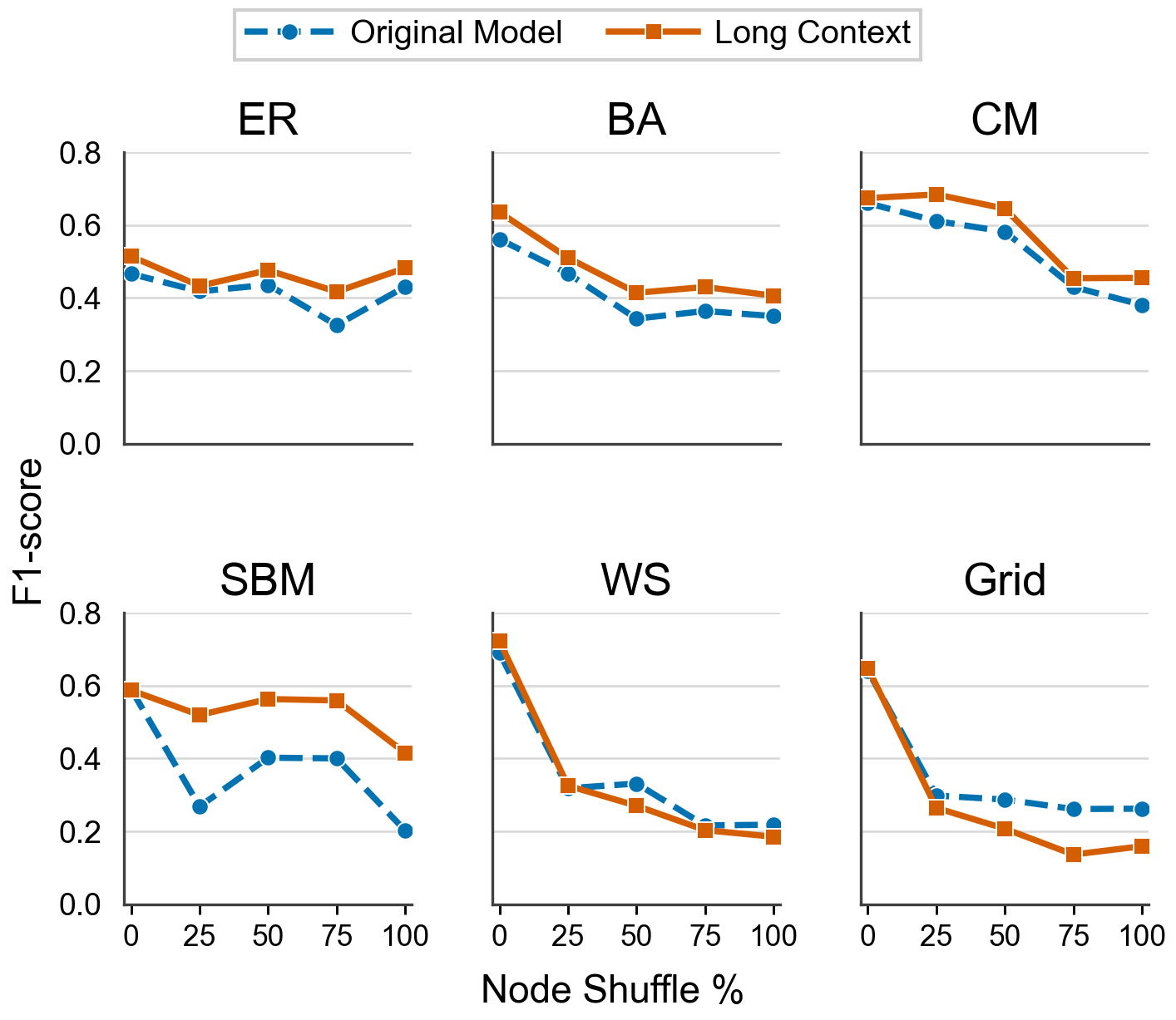}
    \caption{F1-score under node shuffling (L=8 vs. L=128, CC
workload).}
    \label{fig:cc_f1}
\end{figure}

\noindent\textbf{PageRank:}
PR processes every vertex and edge in each iteration. Relabeling changes vertex and neighbor addresses but preserves this full-graph execution schedule. Under complete shuffling, long context retains 57.6\% of its baseline F1-score and improves mean F1-score by 11.0\% over short context.
\begin{figure}
    \centering
    \includegraphics[width=1\linewidth]{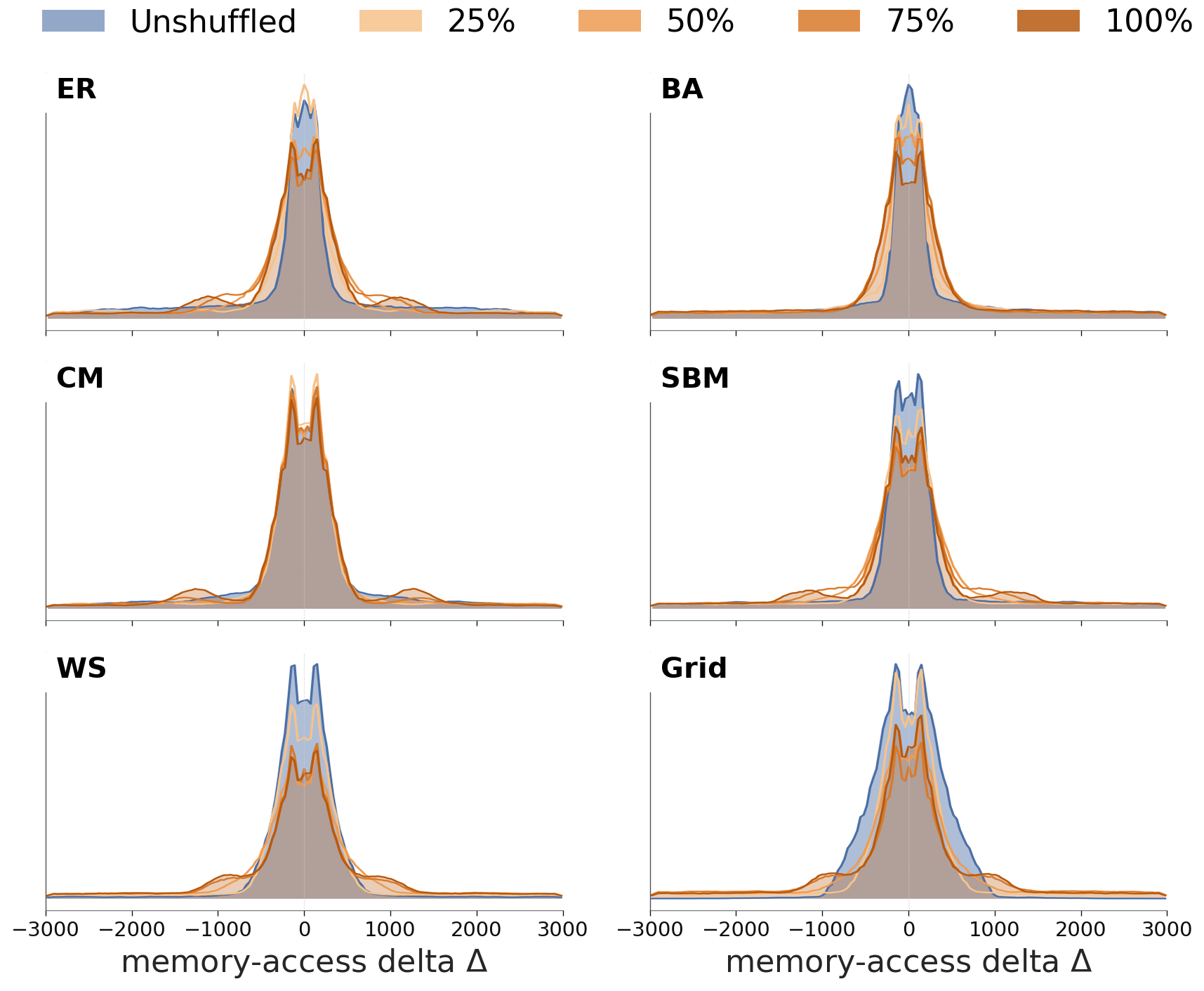}
    \caption{Memory-access delta distributions per graph family, before (blue) and after
(orange) node shuffling for PageRank (PR) workload.}
    \label{fig:pr_rod}
\end{figure}

\begin{figure}
    \centering
    \includegraphics[width=.8\linewidth]{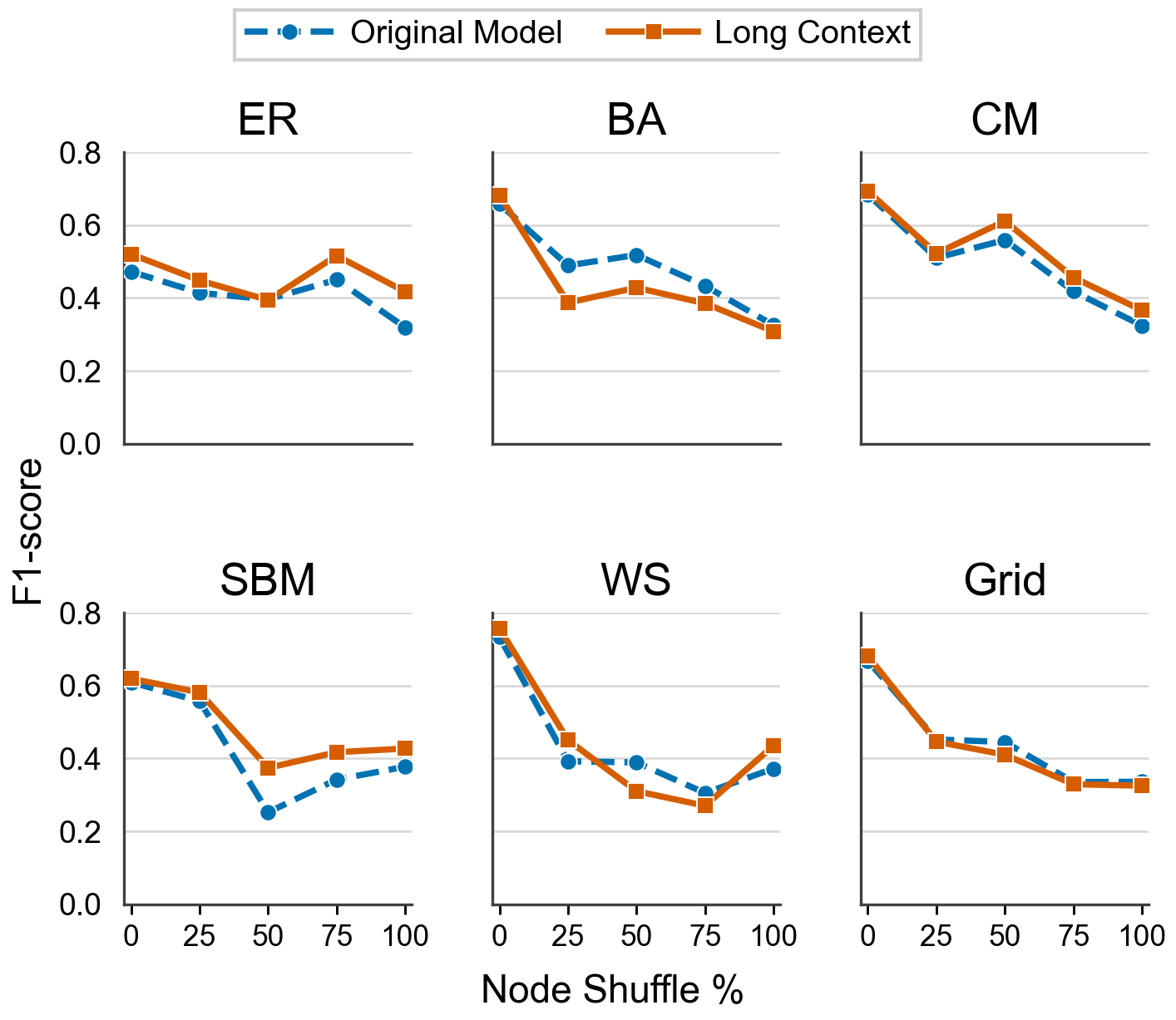}
    \caption{F1-score under node shuffling (L=8 vs. L=128, PR
workload).}
    \label{fig:pr_f1}
\end{figure}

\noindent\textbf{Single-Source Shortest Paths:}
SSSP advances through a frontier generated by conditional distance relaxations. Relabeling remaps temporal address order of frontier accesses, which the marginal delta distribution does not capture. Both context lengths retain only about 30\% of their baseline F1, showing that longer history cannot recover the lost temporal predictability.

}

\subsection{{Real-Graph Validation}}

\begin{figure}
    \centering
    \includegraphics[width=.8\linewidth]{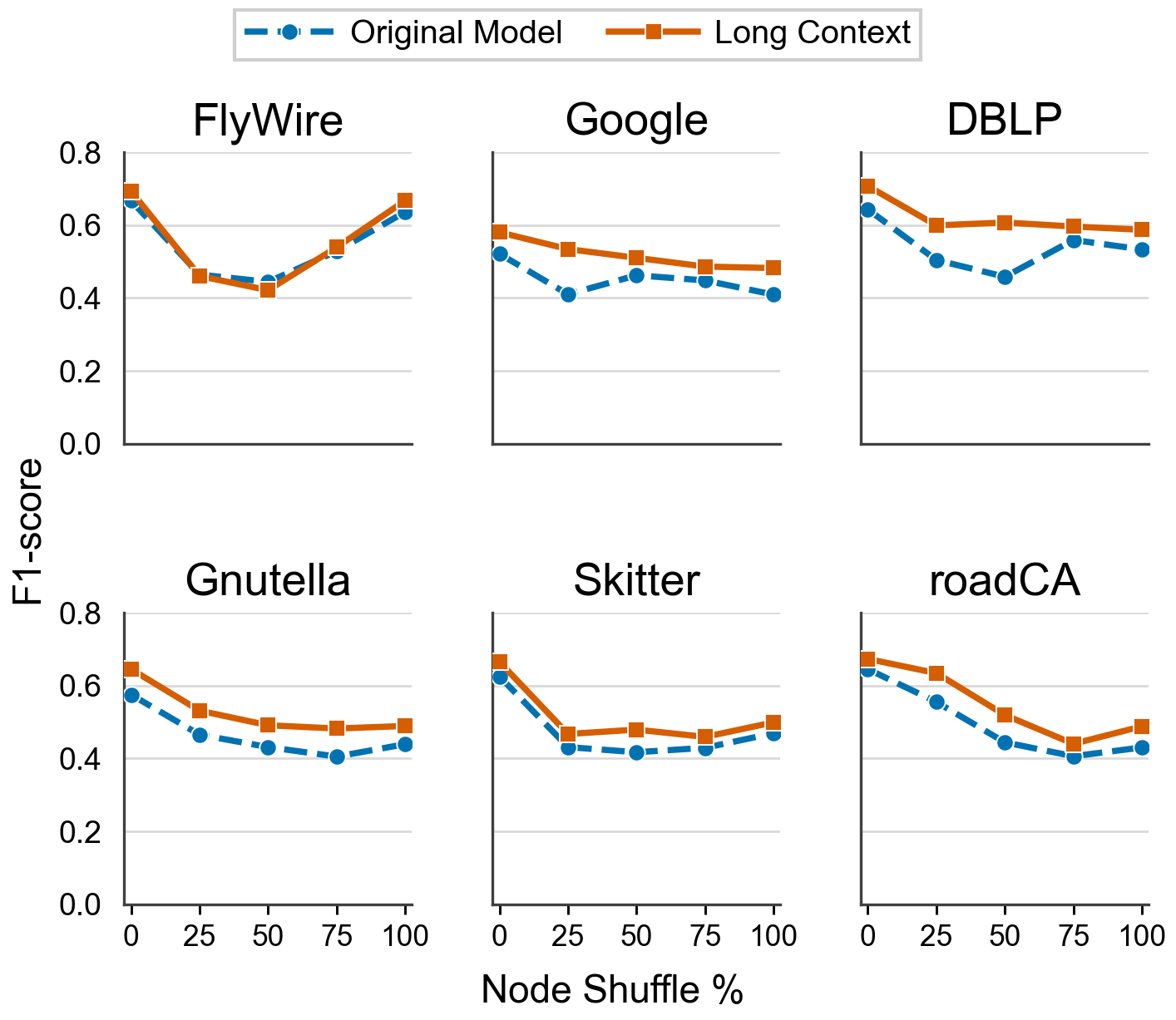}
    \caption{F1-score on six real-world graphs under node shuffling(L=8 vs. L=128, CC
workload).}
    \label{fig:real_cc_f1}
\end{figure}

{
The synthetic results show that random-like native vertex orderings are
more robust to shuffling, and the same trend holds across the six
real-world graphs in Fig.~\ref{fig:real_cc_f1}. For CC, the synthetic results
further show that F1 retention decreases as shuffling moves more deltas
outside the prediction range. The real-graph delta distributions reproduce
this relationship. FlyWire's memory-access delta distribution remains largely unchanged, whereas the
more strongly ordered Skitter and roadCA graphs lose near-zero
concentration and shift more delta mass outside the prediction range,
corresponding to larger F1 degradation. Under complete shuffling, the
long-context model outperforms the short-context model on all six graphs
and improves mean F1-score by 10.2\%.

For CC, long-context F1 retention decreases as
$S_{\mathrm{order}}$ approaches one on both synthetic and real graphs,
showing that structured vertex orderings are more sensitive to
shuffling. FlyWire has a random-like native ordering with $S_{\mathrm{order}}=0.008$ and
retains 96.2\% of its long-context F1-score, whereas roadCA has a
strongly ordered layout with $S_{\mathrm{order}}=0.994$ and retains only
72.5\%. Together, these results show that the relationship among
random-like ordering, range of delta preservation, and F1 robustness transfers
from synthetic to real graphs.

}

\begin{figure}
    \centering
    \includegraphics[width=1.0\linewidth]{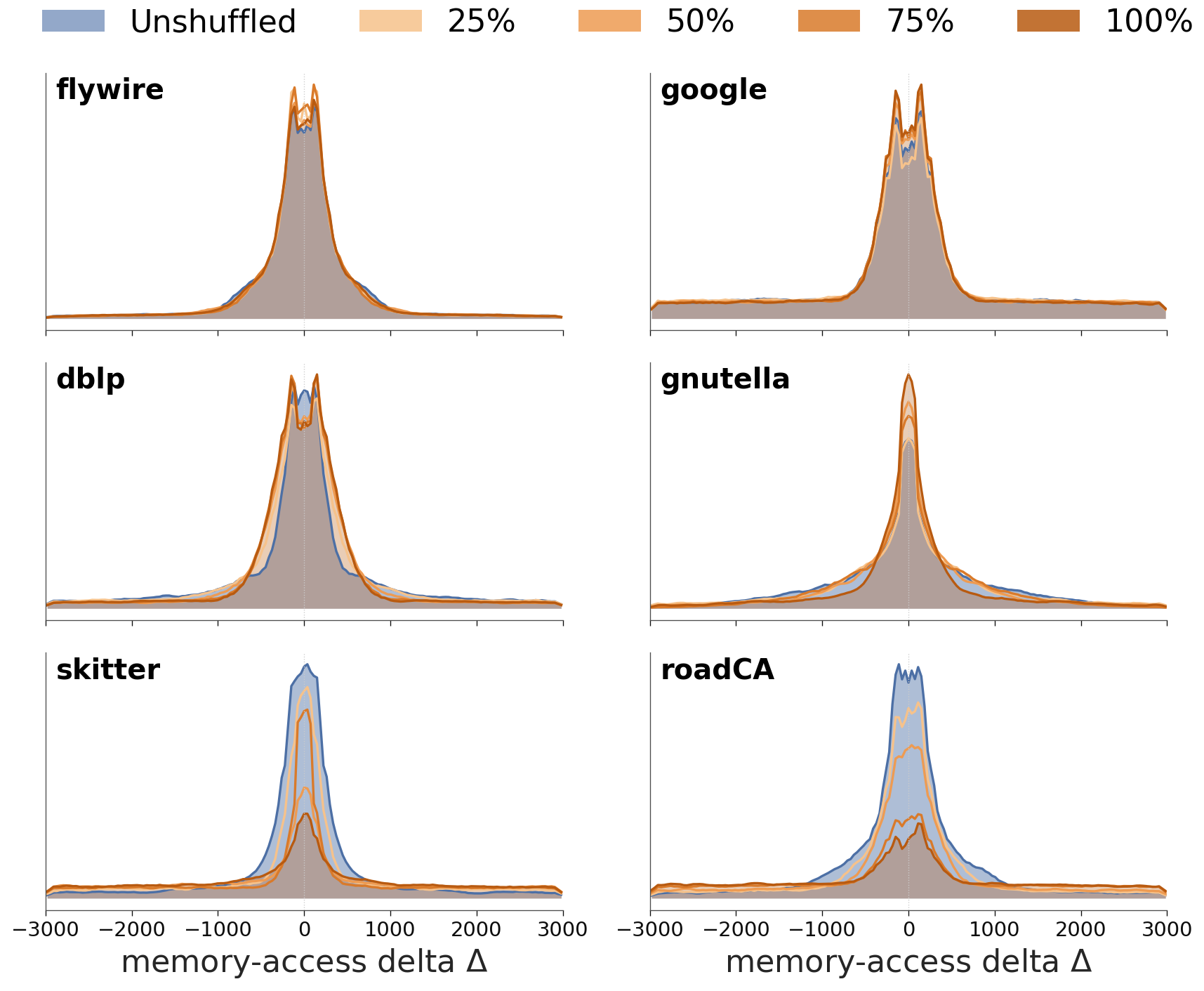}
    \caption{Memory-access delta distributions for CC on six real-world graphs
under progressive node shuffling.}
    \label{fig:real_cc_rod}
\end{figure}

\section{Conclusion}
LoCoMAP decouples look-back length from prediction horizon to study
topology-dependent memory-access prediction across three GAP workloads
and six graph families. Increasing $L$ from 8 to 128 improves F1
by 7.8\% and by up to 36.8\%, while increasing $W$ from 8 to 512 raises
F1 by 88.3\%. Under complete relabeling, long context improves F1
by 33.6\% across ER, BA, CM, and SBM---including a $2.06\times$ gain on
SBM---but not WS or Grid. Same-family and real-graph results show that
robustness emerges when relabeling preserves characteristic delta ranges,
indicating connectivity learning rather than layout memorization. In future work, we
plan to reduce the cost of long context prediction with subquadratic
sequence architectures to enable low-latency ML-based prefetchers.

\section*{Acknowledgment}
This work was supported by the National Science Foundation (NSF) under Award Nos. CSSI-2311870 and OAC-2411446, and by the NSF IDEAS IUCRC under Award No. ASUB00001781.

\textbf{Distribution Statement A:}  Approved for public release. Distribution is unlimited.

\bibliographystyle{IEEEtran}
\bibliography{main}

\end{document}